\newif\iffinal
\finaltrue

\documentclass[sigplan,twocolumn,nonacm,authorversion]{acmart}

\usepackage[normalem]{ulem}

\usepackage{microtype}
\usepackage{xcolor}
\usepackage{url}
\usepackage{filecontents}
\usepackage{tikz}
\usepackage{subfig}
\usepackage{multirow}
\usepackage{booktabs}
\usepackage{graphicx}
\usepackage{xspace}
\usepackage{algorithm}
\usepackage[noend]{algpseudocode}
\usepackage{balance}
\usepackage{amsmath}

\usepackage{amssymb}
\usepackage{amsthm}
\usepackage{amsfonts}
\usepackage{bbm}
\usepackage{threeparttable}
\usepackage{cleveref}
\usepackage{mathtools}
\usepackage{makecell}
\usepackage{graphicx}
\usepackage{lineno}
\usepackage{hyperref}

\renewcommand\footnotetextcopyrightpermission[1]{}
\newcommand{\Sys}{SCServe\xspace}
\newcommand{\Tech}{SLO-customized speculative decoding\xspace}

\newcommand{\ZJ}[1]{\textcolor{red}{ZJ: #1}}
\newcommand{\XM}[1]{\textcolor{blue}{XM: #1}}
\newcommand{\ZL}[1]{\textcolor{cyan}{Zikun: #1}}
\algnewcommand{\LeftComment}[1]{\Statex \(\triangleright\) #1}
\algnewcommand{\LineComment}[1]{\State \textcolor{violet}{\(\triangleright\) #1}}

\iffinal
\else 
\fi

\DeclareMathAlphabet{\mathcal}{OMS}{cmsy}{m}{n}

\usepackage{listings}
\definecolor{mygray}{RGB}{230,230,230}
\theoremstyle{plain}
\newtheorem{theorem}{Theorem}[section]

\newtheorem{lemma}[theorem]{Lemma}

\theoremstyle{definition}
\newtheorem{definition}[theorem]{Definition}

\theoremstyle{remark}

\begin{document}

\date{}

\iffinal

\renewcommand{\Sys}{AdaServe\xspace}

\title{\Sys: SLO-Customized LLM Serving with Fine-Grained Speculative Decoding}

\author{Zikun Li}
\authornote{Contributed equally.}
\affiliation{
  \institution{Carnegie Mellon University}
 \country{USA}
}

\author{Zhuofu Chen}
\authornotemark[1]
\authornote{Work done during internship at Carnegie Mellon University.}
\affiliation{
  \institution{Tongji University}
 \country{China}
}

\author{Remi Delacourt}
\affiliation{
  \institution{EPFL}
  \country{Switzerland}
}

\author{Gabriele Oliaro}
\affiliation{
  \institution{Carnegie Mellon University}
 \country{USA}
}
\author{Zeyu Wang}
\affiliation{
  \institution{Carnegie Mellon University}
 \country{USA}
}
\author{Qinghan Chen}
\affiliation{
  \institution{Carnegie Mellon University}
 \country{USA}
}
\author{Shuhuai Lin}
\affiliation{
  \institution{Carnegie Mellon University}
 \country{USA}
}
\author{April Yang}
\affiliation{
  \institution{Carnegie Mellon University}
 \country{USA}
}
\author{Zhihao Zhang}
\affiliation{
  \institution{Carnegie Mellon University}
 \country{USA}
}
\author{Zhuoming Chen}
\affiliation{
  \institution{Carnegie Mellon University}
 \country{USA}
}
\author{Sean Lai}
\affiliation{
  \institution{Amazon Web Services}
 \country{USA}
}
\author{Xinhao Cheng}
\affiliation{
  \institution{Carnegie Mellon University}
 \country{USA}
}
\author{Xupeng Miao}
\affiliation{
  \institution{Purdue University}
 \country{USA}
}

\author{Zhihao Jia}
\affiliation{
  \institution{Carnegie Mellon University \\Amazon Web Services }
 \country{USA}
}

\settopmatter{printfolios=false, ,printacmref=false}

\fi




\begin{abstract}
Modern large language model (LLM) applications exhibit diverse service-level objectives (SLOs), from low-latency requirements in interactive coding assistants to more relaxed constraints in data wrangling tasks. Existing LLM serving systems, which rely on uniform batching and scheduling strategies, often fail to meet these heterogeneous SLOs concurrently. 
We present \Sys, the first LLM serving system designed to support efficient multi-SLO serving through \textit{\Tech}. \Sys formulates multi-SLO serving as a constrained optimization problem and introduces a hardware-aware algorithm that constructs a speculation tree tailored to each request’s latency target. It features a speculate-select-verify pipeline that enables fine-grained control over decoding speed while maximizing system throughput. 
\Sys further adapts to workload variation by dynamically adjusting speculation parameters. 
Evaluations across diverse workloads show that \Sys reduces SLO violations by up to 4.3$\times$ and improves goodput by up to 1.9$\times$ compared to the best performing baselines, highlighting its effectiveness in multi-SLO serving.

\end{abstract}

\maketitle

\section{Introduction}
\label{sec:intro}

Large language models (LLMs) such as GPT-4, Gemini and Claude have revolutionized various applications including conversational chatbots~\cite{gpt4o,claude35,geminipro, vicuna2023}, code generation tools~\cite{chen2021evaluating, li2022competition,roziere2023code}, and virtual assistants~\cite{vu2024gptvoicetasker, dong2023towards}.
Despite their remarkable capabilities, the deployment of LLMs in practical settings poses significant challenges, particularly in ensuring timely and reliable responses under varying operational conditions.
These applications exhibit varying service-level objectives (SLOs), driven by distinct user expectations and operational contexts.
For example, LLM-powered chatbots must deliver text responses at rates slightly surpassing human reading speeds, approximately 10 tokens per second~\cite{brysbaert2019many,liu2024andes,zhong2024distserve}. In contrast, tools like coding copilots demand much quicker output, tens of tokens in less than 400ms to ensure seamless interactions~\cite{howgithub,mlenhanced}.
Furthermore, recent advances in LLM applications for complex tasks such as reasoning~\cite{jaech2024openai,guo2025deepseek} and data wrangling~\cite{narayan2022can} may accept longer latencies, prioritizing the depth and quality of results over speed.

\begin{figure}
    \centering
    \includegraphics[width=0.92\linewidth]{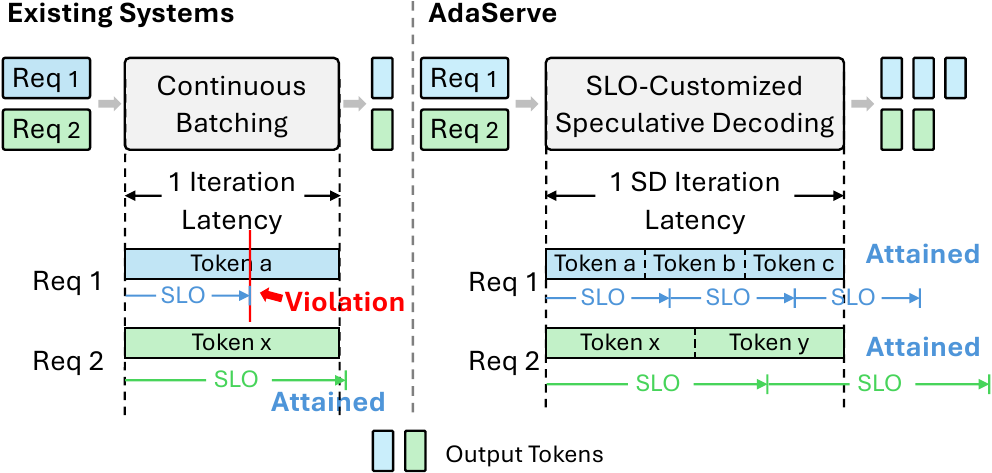}
    \caption{Comparing \Sys and existing systems with continuous batching.}
    \label{fig:motivation}
\end{figure}

The diverse SLOs of various LLM applications present substantial challenges for LLM serving infrastructures.
Existing systems typically employ a \textit{uniform} serving strategy, treating incoming requests homogeneously without considering their specific SLOs.
State-of-the-art systems like vLLM~\cite{vllm} and TensorRT-LLM~\cite{tensorrtllm} leverage \textit{continuous batching} to improve throughput and GPU utilization by batching tokens from different requests~\cite{yu2022orca}. This method schedules execution at the iteration granularity, resulting in uniform per-token latency across batched requests. As shown in \Cref{fig:motivation}, existing systems using continuous batching for multi-SLO LLM serving may violate the stringent SLOs.

Recent work has proposed sophisticated strategies to address the limitations of continuous batching.
For example, Sarathi-Serve~\cite{agrawal2024taming} introduces chunked-prefill, where lengthy prefill requests are partitioned into smaller chunks to expedite the Time-to-First-Token (TTFT).
FastServe~\cite{wu2023fast} employs a preemptive scheduling method to mitigate latency issues caused by long sequences.
VTC~\cite{sheng2024fairness} ensures fair scheduling by tracking the tokens processed for each service and prioritizing those with fewer processed tokens.

Although these attempts improve the adaptivity of serving systems, they lack explicit mechanisms to simultaneously meet diverse and concurrent SLO requirements. They mainly focus on optimizing individual aspects, such as handling requests at different phases or of varying lengths, rather than comprehensively accommodating the multi-SLO demands.

\begin{figure}
    \centering
    \includegraphics[width=0.92\linewidth]{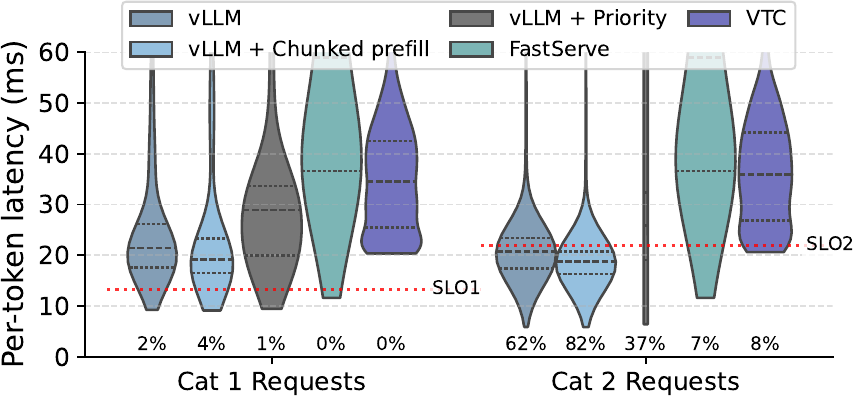}
    \caption{Existing systems cannot efficiently support multi-SLO LLM serving.}
    \label{fig:intro}
\end{figure}

In~\Cref{fig:intro}, we evaluate several serving systems on a multi-SLO workload. As shown, all systems---except for vLLM + Priority---fail to prioritize requests with more stringent SLOs. vLLM + Priority enables such prioritization by allowing urgent requests to preempt non-urgent ones during decoding. However, to meet tight SLOs, the system must constrain batch sizes, which reduces throughput, causes serious congestion, and ultimately degrades overall SLO attainment across all request categories.


Recent research has introduced \textit{speculative decoding} (SD) ~\cite{leviathan2022fast,chen2023accelerating,miao2023specinfer}, a technique designed to reduce the latency of LLM inference by speculatively predicting multiple output tokens at once, albeit with potential inaccuracies. This process is followed by a single verification step using the LLM to simultaneously verify the correctness of the output to ensure lossless generation.
Unlike continuous batching and its derivatives, which conform to the conventional auto-regressive decoding model with its per-token iterative processing, speculative decoding alternates between speculation and verification phases, potentially producing multiple tokens in one step.
This distinct decoding mechanism breaks the intrinsic per-token latency limitations of traditional methods, providing opportunities to dynamically allocate computational resources among batched requests, thereby more effectively meeting the diverse SLO requirements of multiple requests within the same batch.

However, integrating speculative decoding in multi-SLO LLM serving systems presents three key challenges.

\paragraph{Quantifying hardware processing power.} Processing power of modern GPUs significantly influences the maximum number of tokens from all requests that can be verified in parallel, therefore impacting the overall throughput of the serving system. This capacity varies with hardware specifications; however, existing SD methods lack designs optimized for high-throughput serving and often overlook this aspect.

\paragraph{Fine-grained control of decoding speed.}
Existing SD methods generally focus on maximizing decoding speed. However, within the context of multi-SLO serving, the primary objectives are SLO attainment. Instead of maximizing decoding speed for individual requests, it is critical to modulate the decoding rate to use minimal hardware resources while maximally sustaining the SLOs of individual requests, therefore maximizing overall system performance.

\paragraph{Adapting to fluctuating workloads.} Existing SD methods typically adopt a static speculation strategy~\cite{miao2023specinfer}, assuming a fixed workload and uniform performance objectives. However, in multi-SLO serving scenarios, the workload of different applications---as well as the distribution of requests with varying SLO requirements---can change significantly over time~\cite{stojkovic2024dynamollm}.These dynamics alter the optimal tradeoff between speculation aggressiveness and speedup in SD.



To address these challenges, we propose \Sys, the first system designed to support efficient and adaptive multi-SLO LLM serving. \Sys is hardware-aware, utilizing profiling-based roofline models to quantify the available hardware processing power on different GPU platforms.
To fully utilize the hardware capability, we introduce an algorithm that constructs theoretically {\em optimal} draft token trees for all requests. This algorithm ensures that each request is served at the appropriate decoding speed to meet its individual SLO while maximizing overall system throughput.

Building on this foundation, we propose \textit{\Tech}, a practical variant of the optimal algorithm tailored to real-world deployment constraints.
\Tech uses the speculator to estimate the probability of each token being verified by the LLM and constructs a near-optimal token tree for each request based on these estimates.
It adopts a {\em speculate-select-verify} pipeline: the speculator first generates a candidate token tree for each request; \Sys then selects the subset of tokens to verify with the LLM. This decoupling of speculation and selection significantly reduces the overhead of draft model decoding.
Finally, \Sys dynamically tunes the speculation parameters based on the system load, allowing it to smoothly adapt to changes in request distribution and workload intensity over time.

We have conducted extensive evaluations to compare \Sys with existing LLM serving systems across workloads from diverse services and applications. The results show that \Sys consistently outperforms all baselines. Specifically, \Sys achieves up to 4.3$\times$ reduction in SLO violation rate and 1.9$\times$ higher goodput compared to the best-performing baseline. 
Moreover, as the proportion of requests with strict SLOs increases, \Sys maintains high SLO attainment, achieving up to 1.5$\times$ higher SLO satisfaction and 64\% higher goodput relative to the best competing systems.
Finally, when serving requests with strict TPOT SLO requirements, \Sys achieves up to 1.38$\times$ higher goodput than the best baseline systems, demonstrating a significant improvement in the latency-throughput tradeoff.

\section{Background}
\label{sec:background}



\paragraph{LLM serving.} 
Most modern LLMs are based on the Transformer architecture and generate tokens in an \textit{auto-regressive} fashion. In each inference forward pass---referred to as a decoding iteration---the model consumes the entire input sequence and produces a single new token. This newly generated token is then appended to the input sequence for the next iteration.
During each decoding iteration, only one token is produced, yet the entire model must be loaded from device memory. This results in memory-bound execution that under-utilizes GPU's compute resources and motivates batching to promote GPU utilization.
Current LLM serving systems---such as vLLM~\cite{vllm}, TensorRT-LLM~\cite{tensorrtllm} and Sarathi-Serve~\cite{agrawal2024taming}---adopt {\em continuous batching}, which allows sequences to enter and leave the batch at each iteration, further increasing GPU utilization.

However, these systems struggle to support multi-SLO serving with both high SLO attainment and throughput due to two key limitations. First, continuous batching treats all requests uniformly, making it difficult to customize service for individual SLOs. Second, strict latency requirements favor small batch sizes, limiting parallelism and GPU utilization. Conversely, increasing batch size improves throughput but sacrifices latency, reducing the ability to meet tight SLOs.

\begin{figure}
    \centering
    \includegraphics[width=0.9\linewidth]{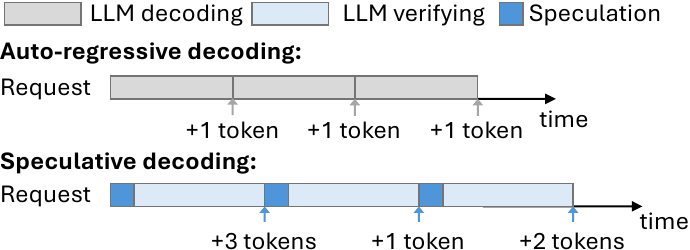}
    \caption{Speculative decoding accelerates LLM inference.}
    \label{fig:spec}
\end{figure}

\begin{figure}
    \centering
    \includegraphics[width=0.82\linewidth]{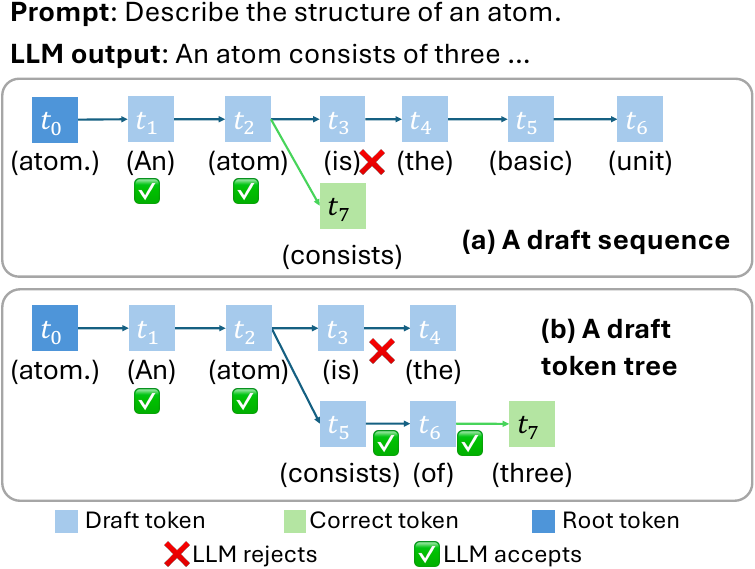}
    \caption{Draft sequence and draft token tree.}
    \label{fig:spec_methods}
\end{figure}

\paragraph{Speculative decoding} 
Speculative decoding (SD) is a technique for accelerating LLM inference by enabling multiple tokens to be generated in a single decoding iteration~\cite{miao2023specinfer, xiaspeculative, leviathan2022fast, chen2023accelerating}. It uses a smaller and faster \textit{draft model} to predict multiple candidate tokens for each request. These candidates are then verified in parallel using the full LLM in a single verification iteration~\cite{miao2023specinfer,eagle1,medusa,fubreak}.

As illustrated in \Cref{fig:spec}, SD consists of two phases: \textit{speculation}, where the draft model proposes token candidates, and \textit{verification}, where the LLM checks their correctness. SD reduces per-token latency by shifting some computation to the smaller model and exploiting the underutilized compute resources of the memory-bound LLM. Verification is performed in parallel and typically incurs minimal additional latency compared to a standard decoding iteration~\cite{chen2024sequoia}.

The output of each speculation phase is a \textit{draft token tree}, rooted at the last generated token (or prompt token if no tokens have been generated). Each node in the tree represents a token, and paths from the root correspond to possible continuation sequences~\cite{miao2023specinfer, chen2024sequoia, medusa, li2024eagle}. The LLM verifies all tokens in the tree in parallel, and the length of the accepted path determines the decoding speedup achieved in that iteration.

\section{Problem Formulation}
\label{sec:problem}

We now formulate the multi-SLO LLM serving problem. In each decoding iteration, given a batch of requests and the token budget---the total number of tokens to verify in this decoding iteration\footnote{The total budget is determined based on hardware profiling. \Sys chooses an optimal budget that balances decoding throughput and latency.}---the goal of multi-SLO serving is twofold: (1) to meet the various TPOT SLO requirements of different requests in the batch and (2) to maximize the number of tokens accepted by the LLM during verification. 

Formally, given a batch of $n$ requests, denoted as $\{r_1, \dots, r_n\}$, and the total token budget $B$, the goal is to construct $n$ token trees $\{T_1, \dots, T_n\}$ for these requests to maximize the expected number of accepted tokens for one decoding iteration, which is expressed as: $E[\sum_{i=1}^n acc(T_i)] = \sum_{i=1}^n E[acc(T_i)]$, where $acc(T)$ is a random variable denoting the number of accepted tokens in $T$ by the LLM verification. This optimization is subject to the following constraints:

\begin{enumerate}
    \item	Budget constraint: The total number of nodes across all token trees must not exceed the hardware budget:
    \begin{align}
        \sum_{i=1}^n |T_i| \le B
    \end{align} where $|T_i|$ is the number of tokens in the $i$-th token tree.
    \item 	TPOT constraint: For each request $r_i$, the expected number of accepted tokens must satisfy the TPOT requirement:
    \begin{align}
    \frac{l_i + t^{spec}}{o_i + acc(T_i)} \le t^{TPOT}_i,\quad \forall  i=1, \dots, n
    \end{align} where $l_i$ denotes the current latency of request $r_i$ starting from the first decoding step, $o_i$ denotes the current number of tokens decoded in request $r_i$, $t^{spec}$ denotes the latency of a decoding iteration and, $t^{TPOT}_i$ denotes the TPOT SLO of request $r_i$.

\end{enumerate}
Intuitively, the budget constraint ensures that the computational intensity of LLM verification stays within the available budget, and the TPOT constraint ensures that the SLO requirements of the requests are satisfied after the current decoding iteration.
For each request $r_i$, we can rewrite the TPOT constraint as: $acc(T_i) \ge (l_i + t^{spec})/t^{TPOT}_i - o_i$.
To further simplify this constraint, we define $A(r_i)=(l_i + t^{spec})/t^{TPOT}_i - o_i$,  which denotes the minimum number of tokens that must be accepted for the $i$-th request in the current decoding iteration to attain its TPOT SLO. With this definition, the TPOT constraint can be simplified as: $acc(T_i) \ge A(r_i), \forall i = 1, \dots, n$.
Since the values of the random variable $acc(T_i)$ is not known during speculation, we relax the TPOT constraint by replacing $acc(T_i)$ with its expectation. 
The relaxed constraint is expressed as:
    \begin{align}
        E[acc(T_i)]  \ge A(r_i), \forall i = 1, \dots, n
    \end{align}
This relaxation not only simplifies the constraint but also enables a more compact expression through the following decomposition of $E[acc(T_i)]$.
\begin{theorem}[Decomposition of the expected number of accepted tokens]\label{thm:decompose}
\begin{align}
    E[acc(T)] = \sum_{v\in T} f(v)
\end{align} where $f(v)$ is the \textit{path probability} of node $v\in T$, defined as the probability in which the LLM accepts the path, which represents a sequence of tokens, from the root node to node $v$ conditioned on the current token sequence of the request.
\end{theorem}

As proven in prior work \cite{chen2024sequoia, li2024eagle}, \Cref{thm:decompose} allows us to rewrite the relaxed TPOT constraint as:
\begin{align} 
    \label{eqn:cons_2_decomposed}
    \sum_{v\in T_i}f(v) \ge A(r_i), \forall i = 1, \dots, n
\end{align}

Based on \Cref{thm:decompose}, we can reformulate the objective of the problem as 
\begin{align}
    \sum_{i=1}^n E[acc(T_i)] = \sum_{v \in \bigcup_{i=1}^n T_i} f(v)
\end{align}

\section{SLO-Customized Serving}
\label{sec:method}

\begin{figure*}[ht]
    \centering
            \includegraphics[width=0.7\linewidth]{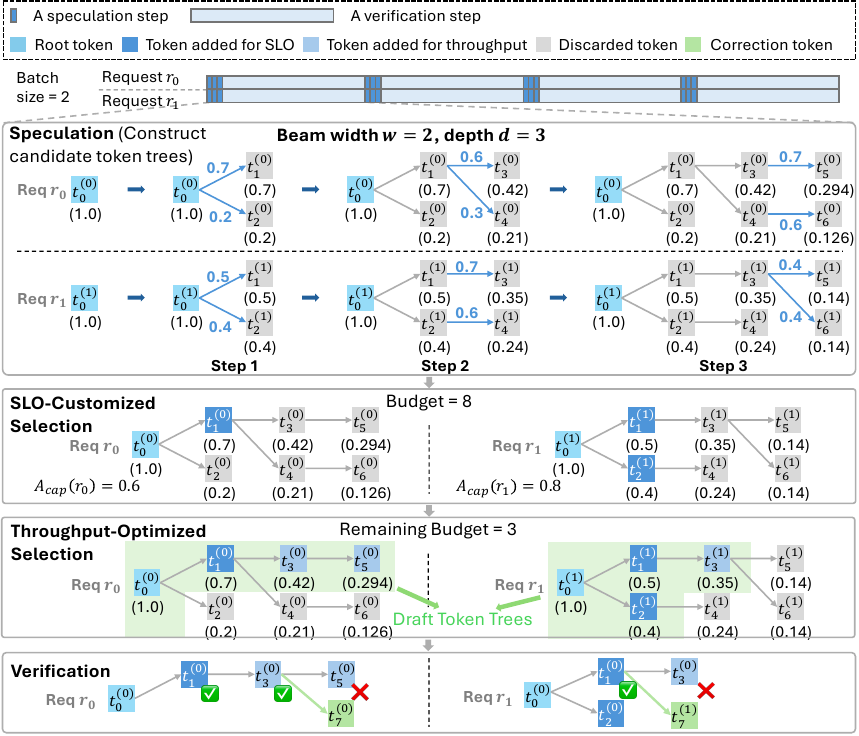}
    \caption{\Tech. In this example, there are two requests in the batch. The budget is 8. In the speculation step, both requests construct a candidate token tree with 3 steps of speculator decoding and beam search where the beam width $w = 2$. 
    During the SLO-customized selection,  $A_{cap}(r_0) = 0.6$, and adding token $t_1^{(0)}$, whose approximated path probability is 0.7, to $T_0$ is enough to attain $r_0$'s TPOT SLO.
    In the same manner, tokens $t_1^{(1)}$ and $t_2^{(1)}$ are added to $T_1$ ($0.5 + 0.4 > 0.8 = A_{cap}(r_1)$).
    This is followed by the throughput-optimized selection with remaining budget 3, where tokens $t_3^{(0)}$, $t_5^{(0)}$ and $t_3^{(1)}$ are added to their corresponding draft token trees for they have the largest approximated path probabilities among the remaining tokens.
Now, \Sys finishes the construction of the draft token trees for both requests. The rest of the tokens in the candidate token trees are discarded.
Finally, the draft token trees are submitted to the LLM for verification.}
    \label{fig:alg}
\end{figure*}

Building on the problem formulation in \Cref{sec:problem}, this section presents our approach to multi-SLO serving. \Cref{subsec:opt_algo} introduces an algorithm that computes a globally \emph{optimal} solution. To make this algorithm practical for real-world LLM serving, we address key integration challenges in \Cref{subsec:challenge}, along with \Sys’s strategies for overcoming them. These strategies are realized in a fine-grained speculative decoding pipeline, detailed in \Cref{subsec:tech}.


\subsection{Optimal Token Tree Construction}
\label{subsec:opt_algo}
We introduce an algorithm that discovers a globally optimal solution to the multi-SLO serving problem, as outlined in \Cref{sec:problem}. 
The algorithm relies on the assumption that the path probability $f(v)$ for any node $v$ in the $T_{inf}(r)$ of request $r$ is known during the construction of the token trees. 
Here, $T_{inf}(r)$ represents the $|V|$-ary infinite-depth token tree for request $r$, where $|V|$ is the vocabulary size.
Each node within $T_{inf}(r)$ corresponds to a token, and the path from the root to any node $v$ forms a sequence of tokens. This tree structure captures all possible output token sequences along with their probabilities (i.e. $f(v)$), which are contingent upon the current token sequence of $r$.

In practice, the assumption of known path probabilities does not always hold; we address such cases in \Cref{subsec:challenge}. Under this assumption, however, we introduce an iterative greedy algorithm to construct optimal token trees in two steps.
In the first step, the algorithm incrementally grows each request’s draft token tree (i.e., $T_i$) by selecting and inserting the node with the highest $f(v)$ from $T_{inf}(r)$.
This procedure is repeated until the TPOT constraints (\Cref{eqn:cons_2_decomposed}) are satisfied for all requests.
If the algorithm determines that the TPOT SLOs cannot be simultaneously met within the given budget, it returns \textsc{INVALID}.
In the second step, the algorithm allocates any remaining budget to insert additional high-$f(v)$ nodes from the union of all $T_{inf}(r_i)$, where each $T_{inf}(r_i)$ represents the $|V|$-ary infinite-depth token tree for request $r_i$.

\Cref{proof:connect} shows that a node chosen greedily by this algorithm is always connected to its parent, ensuring that the constructed token trees are valid. The pseudocode for this algorithm is presented in \Cref{alg:tree_construct_optimal}. 
A formal proof of the algorithm’s optimality is given in \Cref{proof:opt}.


\begin{algorithm}[t]
    \small
    \caption{\label{alg:tree_construct_optimal} An algorithm that outputs the optimal solution to the SLO-aware scheduling problem. }
    {
        \begin{algorithmic}[1]
        \State {\bf Inputs:} requests $\{r_1, \dots, r_n\}$ , a budget $B$ and $f(v)$ for all $v$ in $T_{inf}(r_i), \forall  i = 1, \dots, n$.
        \State {\bf Output:} The optimal draft token tree for each request.

        \State{ $S_{added} \gets \emptyset$}
        \Comment{The set of added nodes.}
        \For{$i = 1, \dots n$}
        \State{Initialize the root of $T_i$.}
        \State{$n_{acc}[i] \gets 1.0$}
        \EndFor
        
        \Comment{Step 1: Add nodes toward SLO requirements.}
        \For{$i = 1, \dots n$}
        \While{$n_{acc}[i] < A(r_i)$}
        \If{$B <= 0$}
        \State{\textbf{Return} INVALID}
        \EndIf
        \State{$v \gets \texttt{GetTop}(T_{inf}(r_i) - S_{added})$}
        \State{$T_i.\texttt{Add}(v)$ }
        \State{$n_{acc}[i] \gets n_{acc}[i] + f(v)$}
       \State{$S_{added}.\texttt{Add}(v)$}
        \State{$B\gets B-1$}
        \EndWhile
        \EndFor

        \Comment{Step 2: Add the rest of tokens.}
        \While{$B \ge 0$}
        \State{$v \gets \texttt{GetTop}(\bigcup_{i=1}^n T_{inf}(r_i) - S_{added})$}
        \State{$i \gets \texttt{GetReqIdx}(v)$}
        \State{$T_i.\texttt{Add}(v)$}
        \State{$S_{added}.\texttt{Add}(v)$}
        \State{$B\gets B-1$.}
        \EndWhile
        \State {\textbf{Return} $\{T_1, \dots, T_n\}$.}
        \end{algorithmic}}
\end{algorithm}

 \subsection{Challenges}
 \label{subsec:challenge}
Applying the optimal token tree construction algorithm in practice presents two key challenges. Next, we describe them and the techniques used in \Sys to address them.
 
\paragraph{Challenge 1: unknown path probabilities $f(v)$.}
\Cref{alg:tree_construct_optimal} assumes that the path probability $f(v)$ for any node $v \in T_{total}$ is known during token tree construction. 
However, in practice, these probabilities are not available a priori. They depend on the LLM's verification of all speculated tokens within the token tree and the subsequent computation of acceptance rates---steps that can only be performed after the token tree has been constructed.

\paragraph{Solution.}
Our key insight is to leverage the logits of the draft model to approximate path probabilities. Specifically, for all $v \in T_{inf}(r_i)$, we approximate:
 \begin{align}
     \prod_{u\in Path(v)}M_q(u|X, Path(u.parent)) \approx  f(v)
 \end{align}
where $M_q$ denotes the draft model used for speculation, which takes a token sequence as input and outputs a probability distribution over the vocabulary. The function $Path(v)$ denotes the sequence of nodes from the root of the token tree to node $v$.
This observation is supported by prior work~\cite{li2024eagle}.

Intuitively, draft models used for speculation are generally trained using the same datasets and with similar objectives as the target LLMs, leading to comparable language modeling capabilities. 
Moreover, recent studies~\cite{zhou2023distillspec, eagle1} show that draft models distilled from large models perform well in speculative decoding. 
Distillation aligns the logits of the draft model closely with those of the large model, making them well-suited for approximating conditional acceptance probabilities. 
Consequently, the logits of the draft model are accurate surrogates for estimating $f(v)$ during token tree construction.

\paragraph{Challenge 2: high speculation overhead.}
In speculative decoding, the draft model generates output tokens in an auto-regressive manner, introducing significant speculation overhead.
In \Cref{alg:tree_construct_optimal}, both construction steps rely on the \texttt{GetTop} operation, which selects the node with the highest path probability from one or multiple token trees.
For a single token tree, a straightforward implementation of \texttt{GetTop} maintains a global candidate set containing all nodes whose parents have already been processed by the draft model but which themselves have not yet been decoded. Each candidate node is associated with an approximated path probability.

The candidate set is initialized with the root node of the token tree, assigned a path probability of 1.
\Cref{alg:tree_construct_optimal} then repeatedly selects the node with the highest path probability from the candidate set and adds it to the token tree.
Once a node is decoded by the draft model, its child nodes, along with their approximated path probabilities, are inserted into the candidate set. 
The second step of \Cref{alg:tree_construct_optimal} follows a similar strategy.

However, this approach results in $(B - n)$ draft model decoding steps, where $B$ is the total token budget and $n$ is the number of requests in a batch. Since each new node addition requires a draft model decoding, and $B \gg n$ in practical settings, the cumulative speculation overhead becomes prohibitively large.

\paragraph{Solution.} 
The inefficiency in \Cref{alg:tree_construct_optimal} arises from the interleaving of top-node selection and draft model decoding, where each decoding step processes only one token. 
To address this issue, we decouple token tree construction into two distinct phases: a {\em speculation phase} and a {\em selection phase}.

In the speculation phase, we use parallel decoding to construct a candidate token tree sufficiently large to cover all potential top nodes. In the subsequent selection phase, we identify the highest-probability nodes from the candidate tree to construct the final token trees for LLM verification. 

This separation of the speculation and selection phases eliminates the inefficiency of interleaved decoding and selection, allowing the draft model to operator more efficiently. The soundness of this method is supported by the following theorem.

\begin{theorem}[Bounding the optimal draft token tree]\label{thm:bound}
Let the total token budget be $B$ and let $T_{opt}$ denote the optimal draft token tree produced by  \Cref{alg:tree_construct_optimal}. Let $D_{opt} = D(T_{opt})$ be the maximum depth of any node in $T_{opt}$. $T_{opt}$ is guaranteed to be a subtree of a candidate tree $T_{cand}$ constructed via a $D_{opt}$-step beam search with beam width $B$.
\end{theorem}

\Cref{thm:bound} implies that in the speculation phase, a candidate tree containing $T_{opt}$ can be constructed using only $D_{opt}$ decoding steps from the draft model via beam search. Generalizing this result, the optimal token trees for all requests can be covered using at most $D_{opt} = \max(D(T_{opt}(r_i))$, where $i = 1, \dots, n$, representing the number of decoding steps required.

Furthermore, if $\text{argmax}_{i=1}^n(D(T_{opt}(r_i)) = j$, we can derive: $D_{opt} = D(T_{opt}(r_j)) \le |T_{opt}(r_j) - 1|  \le \sum_{i=1}^n |T_{opt}(r_i) - 1| = \sum_{i=1}^n |T_{opt}(r_i)| - n = B - n$.
Equality holds only in rare cases where all but one optimal token tree consist solely of root nodes, while the remaining tree forms a long sequence. In practice, such extreme imbalance is unlikely to occur, and empirically, we observe that $D_{opt} \ll B-n$.

Importantly, it is not necessary to include all tokens from $T_{opt}$, particularly when doing so would incur high decoding costs. By tuning the beam search depth $d$ and beam width $w$, \Sys allows a flexible trade-off between speculation accuracy and decoding overhead. This separation of speculation and selection phases significantly improves the efficiency of speculator decoding by leveraging parallelism. Based on these insights, we propose \textit{\Tech} as the core technique of \Sys.

\subsection{SLO-Customized Speculative Decoding} 
\label{subsec:tech}
Each decoding iteration in \Tech consists of four steps: speculation, SLO-customized selection, throughput-optimized selection, and verification. This section introduces the design and purpose of each stage. The pseudocode for these steps is presented in \Cref{alg:tree_construct_emp}.

\paragraph{Step 1: speculation.}
In the speculation step, a beam search algorithm is used to construct candidate token trees for each request, as illustrated in \Cref{fig:alg}. 
Initially, each request’s candidate token tree consists solely of a root node, which represents the last generated token or the prompt if no text has yet been generated. 
The $n$ root tokens for all requests are processed in parallel.
In the first decoding step, the draft model processes all root nodes and produces $|V|$ potential child nodes for each node. 
For each request, the $w$ child nodes with the highest approximated path probabilities $M_q(v|X, Path(v.parent))$ are selected and added to its candidate token tree.

Starting from the second decoding step, the draft model processes all tokens selected in the previous step---$n \times w$ tokens in total---in parallel. 
For each request, the draft model generates $w \times |V|$ potential tokens, and the $w$ with the highest approximated path probabilities are chosen to expand the candidate token tree further. 
After completing $d$ speculation steps, each request $r_i$ has an associated candidate token tree $T_{cand}(r_i)$ with a depth of $d$, where all layers except the first contain exactly $w$ nodes. 

An example is shown in \Cref{fig:alg}, where the draft model performs three decoding steps to construct candidate token trees with a depth of 3 and a beam width of 2. 
The parameters $d$ and $w$ are dynamically determined based on the system load (see \Cref{sec:system}).

The speculation phase is followed by two selection phases: the SLO-customized token selection and the throughput-optimized token selection.

\paragraph{Step 2: SLO-customized token selection.}
In this phase, each request selects tokens from its candidate token tree to construct a draft token tree that satisfies its TPOT requirement. 
According to the TPOT constraint (\Cref{eqn:cons_2_decomposed}), the total approximated path probabilities of all nodes in a request's draft token tree must exceed $A(r)$, the minimum number of tokens that must be accepted to attain the SLO.

However, this requirement may not always be feasible. The number of verifiable tokens per request is upper bounded by $d + 1$. If $A(r) > d + 1$, the SLO cannot be fully satisfied within the current iteration. In this case, \Sys caps the target threshold using $A_{cap}(r) = \min(A(r), d + 1)$, indicating the maximum attainable progress toward the SLO for require $r$.
For each request $r$, \Sys iteratively selects nodes from $T_{cand}(r_i)$ with the highest approximated path probabilities and adds them to the draft token tree $T_i$ until the cumulative approximated path probabilities of all tokens in $T_i$ reaches or exceeds $A_{cap}(r_i)$. 

As shown in the SLO-customized selection step of \Cref{fig:alg}, request $r_0$ requires $A_{cap}(r_0) = 0.6$, so only node $t_1^{(0)}$ is added to $T_0$. For request $r_1$, $t_1^{(1)}$ alone is insufficient, so $t_2^{(1)}$ is also added to satisfy $A_{cap}(r_1) = 0.8$.

When the total budget is insufficient to meet all SLOs, \Sys prioritizes slower requests---those with larger $A(r_i)$---by processing them in descending order of their SLO requirement.
However, challenges arise when satisfying $A_{cap}(r_i)$ for request $r_i$ requires selecting a large number of low-probability nodes. This leads to diminishing returns and may deplete the budget disproportionately. 
In extreme cases, all nodes in $T_{cand}(r_i)$ might be added to $T_i$ without meeting the threshold, monopolizing the budget and degrading system-wide performance.

To address this issue, \Sys enforces a per-request token limit $n_{max}$ during the SLO-customized selection phase. This constraint prevents excessive allocation to low-probability nodes and ensures more balanced and efficient use of recourses across all requests. 

\paragraph{Step 3: throughput-optimized selection.}
While the first two phases focus on satisfying the SLOs of individual requests, this phase 
aims to maximize overall system throughput. \Sys selects the remaining tokens by globally ranking all candidate nodes across requests based on their approximated path probabilities and greedily adding the top-scoring nodes to the draft token trees. This process continues until the overall token budget is exhausted.

As illustrated in the throughput-optimized token selection step of \Cref{fig:alg}, suppose the remaining budget is 3. \Sys selects the top three nodes---$t_3^{(0)}$, $t_3^{(1)}$, and $t_5^{(0)}$---as they have the highest approximated path probabilities among all remaining candidate nodes, and sequentially adds them to the corresponding draft token trees.

\paragraph{Step 4: verification.}
In the final step, \Sys submits the draft token trees for all requests to the LLM, which verifies the correctness of all speculated tokens in parallel. \Sys adopts a tree-based verification strategy, as introduced in prior work~\cite{miao2023specinfer, chen2024sequoia, sun2024spectr, leviathan2022fast}, which efficiently verifies multiple speculative paths by leveraging shared prefixes and minimizing redundant computation. This parallel verification step determines which tokens are accepted and enables the system to advance the decoding process accordingly.

\begin{algorithm}[t]
    \small
    \caption{\label{alg:tree_construct_emp}\Tech: an adaption of \Cref{alg:tree_construct_optimal} that addresses real-system challenges. 
    }
    {
        \begin{algorithmic}[1]
        \State {\bf Inputs:} a small model $M_q$, requests $\{r_1, \dots, r_n\}$, a budget $B$, depth $d$, beam width $w$ and $n_{max}$ ,the upper limit of tokens added to a request's draft token tree during SLO-customized selection.
        \State {\bf Output:} The token tree for each request.

\Comment{Initialization.}
\State{ $S_{added} \gets \emptyset$}
\Comment{The set of added nodes.}
\For{$i = 1, \dots n$}
\State{Initialize the root of $T(r_i)$.}
\State{$n_{acc}[i] \gets 1.0$}
\State{$B\gets B-1$.}

\EndFor

\Comment{The speculation phase.}
\State{$\{T_{cand}(r_1), \dots, T_{cand}(r_n)\} \gets \texttt{Spec}(M_q, \{r_1, \dots, r_n\}, d, w)$}

    \Comment{SLO-customized selection.}
    \State{$\{r_1', \dots, r_n'\} = \texttt{Sort}(\{r_1, \dots, r_n\}, \texttt{key}=A(r))$}
    \State{$n_{acc}' = \texttt{Sort}(n_{acc}, 
 \texttt{key}=A(r))$}

    \For{$i = 1, \dots n$}
    \While{$n_{acc}'[i] < A_{cap}(r_i') \wedge |T(r_i')| < n_{max}\wedge  B\ge 0$}
    \State{$v\gets \texttt{GetTop}(T_{cand}(r_i') - S_{added})$}
    \State{$T(r_i').\texttt{Add}(v)$}
    \State{$n_{acc}'[i] \gets n_{acc}'[i] + M_q(v|X(r_i'), Path(v.parent))$}
    \State{$S_{added}.\texttt{Add}(v)$}
    \State{$B\gets B-1$.}
    \EndWhile
    \EndFor

    \Comment{Throughput-optimized selection.}
    \While{$B \ge 0$}
        \State{$v \gets \texttt{GetTop}(\bigcup_{i=1}^n T_{cand}(r_i) - S_{added})$}
        \State{$r \gets \texttt{GetReq}(v)$}
        \State{$r.T.\texttt{Add}(v)$}
        \State{$S_{added}.\texttt{Add}(v)$}
        \State{$B\gets B-1$.}
    \EndWhile
    \State {\textbf{Return} $\{T(r_1), \dots, T(r_n)\}$.}
        \end{algorithmic}}
\end{algorithm}
\section{System Design and Optimizations}\label{sec:system}

\subsection{Overview of \Sys}

\begin{figure}[t]
    \centering
    \includegraphics[width=0.92\linewidth]{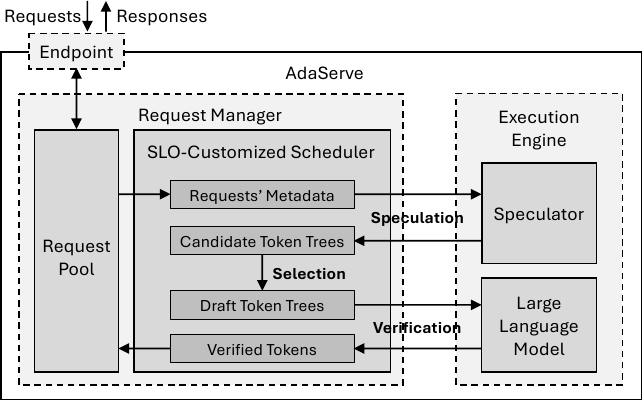}
    \caption{Overview of \Sys.}
    \label{fig:overview}
\end{figure}

\Cref{fig:overview} presents an overview of \Sys, which consists of two main components: the \textit{request manager} and the \textit{execution engine}.
The request manager maintains a pool of active requests and includes an SLO-customized scheduler that implements \Tech. The execution engine is responsible for executing both the draft and target  models on GPUs.
At the beginning of each speculation iteration, the SLO-customized scheduler retrieves all active requests from the request pool and initiates the speculation phase of \Tech by instructing the execution engine to run the draft model for $d$ decoding steps. Once the speculation phase completes, the selection phases are executed to construct draft token trees for all requests.
These draft token trees are then submitted to the large language model for verification. After verification, the logits of the nodes in each tree are returned to the SLO-customized scheduler. The scheduler uses these logits to identify the verified tokens for each request, which are then stored back into the request pool for the next iteration or final output assembly.

\subsection{System Optimizations}

\paragraph{Adaptive control.}
The depth ($d$) and beam width ($w$) of the speculation tree directly affect the decoding overhead of the draft model. Larger values of $d$ and $w$ can significantly increase speculation cost, especially under high system load. In practice, the number of active requests $n$ varies over time, and using fixed values for $d$ and $w$ fails to adapt to this dynamic workload.

When many requests are active, the average token budget per request decreases, limiting the viable depth and width of each token tree. In such cases, large $d$ and $w$ values generate excessive speculative tokens that are likely to be discarded, leading to wasted computation. Conversely, when the system load is low, each request can be allocated more tokens. Using small fixed values in these cases limits the potential performance gains from deeper and wider trees.

To address this issue, \Sys dynamically adjusts $d$ and $w$ based on the current number of active requests $n$ using the following policy at the begin of each iteartion:
\begin{align}
    d & = \text{clip}(D_{max}, D_{min}, \lfloor\frac{B_1}{n + c_1}\rfloor-1)\\
    w &= \text{clip}(W_{max}, 1, \lfloor\frac{B_2}{n}\rfloor +c_2)
\end{align}
Here, $D_{max}$, $D_{min}$, and $W_{max}$ are predefined bounds for tree depth and width. $B_1$ and $B_2$ denote the total number of tokens allocated per decoding step for the verifier and the speculator, respectively. $c_1$ and $c_2$ are tunable constants, selected via grid search. The clip function constrains its third argument within the specified upper and lower bounds.

Speculation depth has the most significant impact on overhead. The formula for $d$ is designed to ensure that the number of speculative tokens remains within the average verification budget per request, minimizing the likelihood of excessive speculative computation being wasted.

\paragraph{GPU optimizations.}

Enabling efficient multi-SLO serving on GPUs introduces additional challenges. One such challenge involves leveraging CUDA graphs~\cite{cudagraph}, which reduce kernel launch overhead by capturing a sequence of GPU kernel executions and their dependencies into a computation graph. This graph can then be replayed efficiently in subsequent iterations. However, reusing a CUDA graph requires that kernel shapes and input dimensions remain identical to those used during the initial capture.
\Sys utilizes CUDA graphs to accelerate draft model decoding. In the speculation phase, decoding steps from the second to the $d$-th step perform the same operations: each of the $n$ requests generates $w$ tokens, resulting in consistent computation patterns. Furthermore, across iterations with the same number of active requests $n$, the decoding shapes and workloads remain unchanged. This structural regularity allows \Sys to reuse pre-captured CUDA graphs across multiple steps and iterations, significantly reducing GPU launch overhead.

\section{Evaluation}
\label{sec:eval}

\subsection{Experimental Setup}
\paragraph{Implementation and device}
We implement \Sys on top of FlexFlow Serve \cite{jia2019flexflow}, a low-latency, high-throughput LLM serving framework. To further optimize performance, we integrate the batched prefill kernel from FlashInfer~\cite{flashinfer}, a high-performance kernel library for LLM serving. This kernel is adapted for both speculation steps and LLM verification.
All evaluations are performed on a compute node equipped with four NVIDIA A100 80GB GPUs, interconnected via NVLink. The node is powered by an AMD EPYC 7763 CPU with 64 cores (128 threads) and 256 GB of DRAM.

\begin{table}
    \small
    \centering
    \begin{tabular}{ccc}
    \Xhline{2\arrayrulewidth}
        \textbf{Model} & \textbf{Parallelism} & \textbf{GPUs}\\
        \hline
        Llama3.1-70B-Instruct & 4-way TP & 4 $\times$ A100 80G \\
        Qwen2.5-32B-Instruct &  2-way TP & 2 $\times$ A100 80G\\
        \Xhline{2\arrayrulewidth}
    \end{tabular}
    \caption{Evaluation setups for different models. "TP" stands for tensor parallelism. }
    \label{tab:setup}
\end{table}

\paragraph{Models}
\Cref{tab:setup} summarizes the models, parallelism strategies, and GPU configurations used in our evaluation. This setup is applied consistently across \Sys and all baseline systems. For speculative decoding experiments, the draft model is colocated with the base model on one of the GPUs.
We use Llama3 \cite{llama3} and Qwen2.5 \cite{yang2024qwen2} models, as their architectures are representative of modern LLMs.
The draft model is selected as the smallest available model within the same family as the corresponding base model: LLaMA-3.2-1B-Instruct is used for LLaMA 3, and Qwen2.5-0.5B-Instruct for Qwen2.5.

\paragraph{Baselines.}
We compare \Sys against state-of-the-art LLM serving systems, including vLLM~\cite{vllm}, Sarathi-Serve~\cite{agrawal2024taming} and vLLM augmented with speculative decoding. 
vLLM introduces PagedAttention~\cite{vllm}, a memory management technique that improves throughput by mitigating fragmentation.
Sarathi-Serve~\cite{agrawal2024taming} leverages chunked prefill to jointly batch the prefill and decoding stages across multiple requests, enhancing hardware utilization and reducing per-token latency.
We also evaluate speculative decoding baselines built on top of vLLM, which implement efficient sequence-based speculative decoding.
We include variants with different speculation lengths, denoted as vLLM-Spec($n$), where $n$ represents the number of speculated tokens. All evaluations use the latest version of vLLM available at the time of submission (v0.8.4).
\begin{table}
    \centering
    \small
    \begin{tabular}{c|ccc} 

        \Xhline{2\arrayrulewidth}
         \textbf{Category}&  {\bf \begin{tabular}{{@{}c@{}}}Cat. 1\\\end{tabular}} &  {\bf \begin{tabular}{{@{}c@{}}}Cat. 2\\\end{tabular}}& {\bf \begin{tabular}{{@{}c@{}}}Cat. 3\\\end{tabular}}\\ 
        \hline
         \textbf{App}&  Coding copilot&  Chatbot& Summarization\\ 
         \textbf{SLO }&  1.2 $\times$ Baseline latency &  50ms & 150ms \\ 
        \Xhline{2\arrayrulewidth}
    \end{tabular}
    \caption{Request categories and their SLOs. }
    \label{tab:slo}
\end{table}

\begin{figure}
        \centering
        \includegraphics[width=0.98\linewidth]{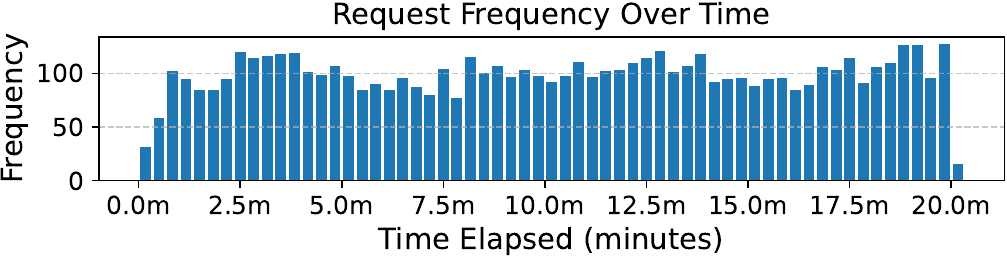}
        \caption{Request frequency of the real-world trace.}
        \label{fig:trace_freq}
\end{figure}

\begin{figure*}[ht]
    \centering
    \begin{minipage}{\textwidth}
        \centering
        \includegraphics[width=\textwidth]{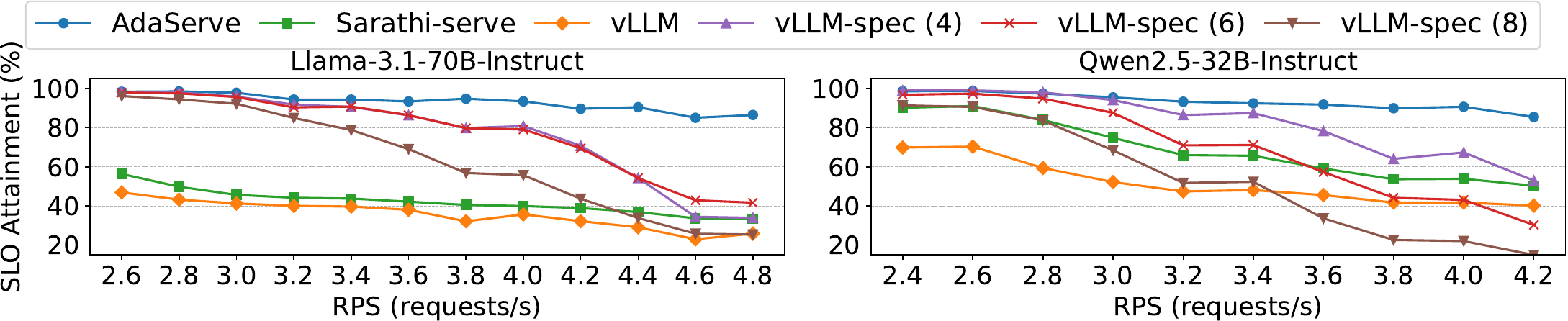}
        \vspace{-2em}
        \caption{SLO attainment w.r.t. RPS.}
        \label{fig:rps_attainment}
    \end{minipage}
    \begin{minipage}{\textwidth}
        \centering
        \includegraphics[width=\textwidth]{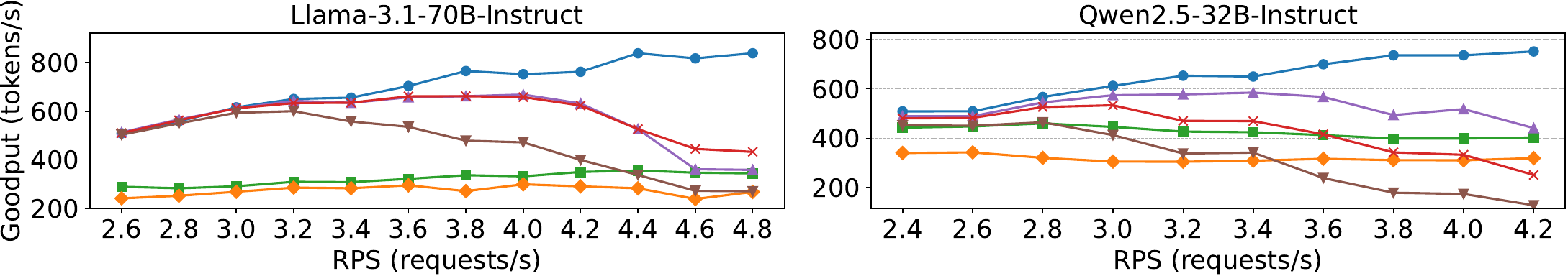}
        \vspace{-2em}
        \caption{Goodput w.r.t. RPS.}
        \label{fig:rps_goodput}
    \end{minipage}
\end{figure*}

\begin{figure*}[t]
    \centering
    \includegraphics[width=\textwidth]{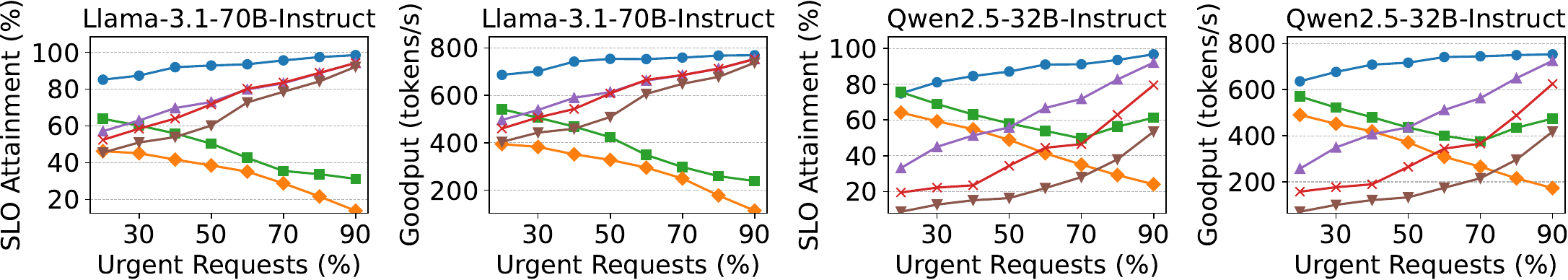}
    \vspace{-1em}
    \caption{SLO attainment and goodput w.r.t. urgent request proportion.}
    \label{fig:proportion}
\end{figure*}

\begin{figure*}[t]
    \centering
    \includegraphics[width=\textwidth]{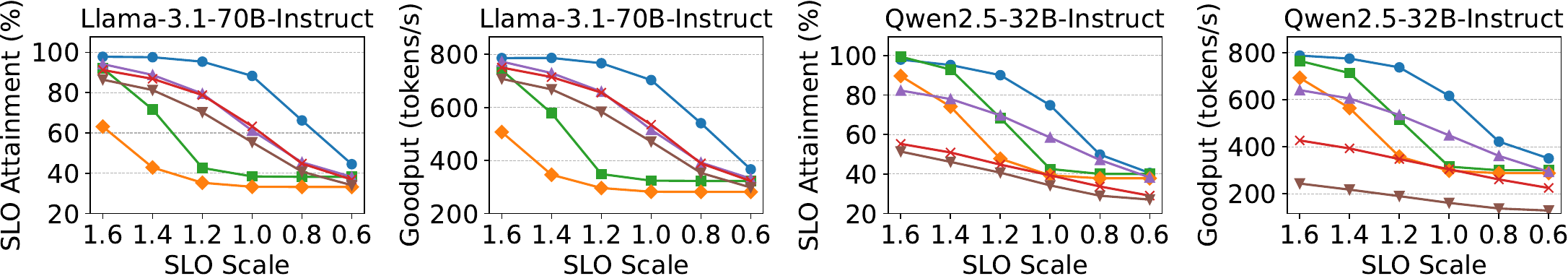}
    \vspace{-1em}
    \caption{SLO attainment and goodput w.r.t. SLO scale.}
    \label{fig:slo_scale}
\end{figure*}

\paragraph{Workloads.}
We evaluate \Sys using a mixture of requests from different applications, each with distinct SLO requirements, following prior work~\cite{zhong2024distserve}.
We consider requests from three categories, as summarized in \Cref{tab:slo}.
For each model, we measure a \textit{baseline latency} when the system load is close to zero, which serves as a reference point for setting TPOT SLOs across different request categories following prior work \cite{zhong2024distserve,li2023alpaserve}.

For this category, we simulate code completion tasks using prompts from the HumanEval dataset~\cite{chen2021evaluating}, which contains 164 programming problems.
The SLO for this category is set to 1.2$\times$ the baseline latency, a stringent target that permits a 20\% slowdown to support high-throughput serving.
This SLO setting aligns with the SLO for latency-sensitive interactive applications in MLPerf v5.0~\cite{reddi2020mlperf}, which specifies 40ms per token for Llama 70B models~\cite{mlperf5,mlperfdatacenter}.

The second category includes chatbot requests.
To maintain a responsive user experience, chatbots must stream tokens faster than users can consume them.
While normal human reading speed is 200-300 words per minute, skimming can occur at 2-4$\times$ that rate, translating to a per-token latency requirement of slightly under 50ms~\cite{rayner2016so}.
Thus, we adopt 50ms per token as the SLO for this category.
We use the Alpaca dataset \cite{alpaca} which contains 52k instruction-following examples to simulate chatbot interactions.

The third category includes tasks with relaxed latency requirements, such as  LLM-based summarization, where higher TPOT SLOs are acceptable. 
We set the SLO to 150ms per token, consistent with prior work and benchmark settings~\cite{zhong2024distserve,mlperf5,mlperfdatacenter}.
For this category, we use summarization tasks from the CNN/DailyMail dataset~\cite{bai2023longbench}, which contains news articles paired with human-written summaries. 

We use the timestamps from a real-world trace from previous work, visualized in~\Cref{fig:trace_freq}, to generate traces in our evaluation~\cite{patel2024splitwise}. We truncate and rescale the trace to obtain traces with different averaged request per second (RPS).
For each arriving request, we first sample its category according to a specified probability distribution and then sample a request from the dataset uniformly.

\paragraph{Metrics.}
We use \textit{SLO attainment} and \textit{goodput} as our primary metrics. SLO attainment is the percentage of requests in a workload that meet their SLO. Specifically, a request is considered to fulfill its SLO if its average per-token latency is no greater than the specified TPOT SLO threshold. Goodput is measured as the number of tokens generated per second for requests that successfully attain their SLO.

\subsection{End-to-End Comparison}
\label{subsec:e2e}

\paragraph{Changing request arrival rate}
We first evaluate the end-to-end performance of \Sys under increasing request arrival rates by comparing \Sys's SLO attainment and goodput against those of vLLM, Sarathi-Serve, and vLLM-Spec.
The workload consists of 60\% category 1 requests, 20\% category 2 requests, and 20\% category 3 requests. This mix represents a peak load scenario for latency-critical tasks (category 1), while workloads for categories two and three are lighter, allowing us to assess system performance under stringent task conditions.

As shown in \Cref{fig:rps_attainment} and \Cref{fig:rps_goodput}, \Sys consistently achieves higher SLO attainment and goodput across all models and request rates compared to the baselines, with the performance gap widening as the request rate (RPS) increases. 
\Sys improves the SLO attainment by 2.1$\times$ and 1.6$\times$ over the best baseline on the two models, respectively.
At the highest RPS, \Sys reduces the number of unattained requests by 4.3$\times$ and 3.2$\times$, respectively.
In terms of goodput, \Sys delivers 1.9$\times$ and 1.7$\times$ higher goodput than the best baseline under the two settings.

vLLM and Sarathi-Serve both struggle to meet stringent SLOs. This is primarily due to their reliance on continuous batching, which enforces a uniform TPOT SLO across all requests in a batch. 
As the request rate increases, the running batch size also increases, leading to higher per-token latency and lower SLO attainment. 
In contrast, \Tech enables \Sys to dynamically allocate hardware resources based on individual request SLOs, allowing it to prioritize latency-critical requests. 
This selective prioritization leads to significantly improved SLO attainment and goodput, even with high request arrival rates.


vLLM-Spec outperforms other baselines; however, its performance degrades significantly as the request arrival rate increases.
These results highlight the limitations of static speculation methods, wich fail to account for diverse SLO requirements and dynamic workload variants.
Specifically, vLLM-Spec adopts a fixed speculation strategy that cannot adapt to the applications' latency needs or the system's current workload.
When the workload is low, allocating only a small number of speculative tokens results in under-utilization of hardware and limited performance gains.
Conversely, under high-load conditions with large batch sizes, the static strategy generates too many speculated tokens, leading to high verification overhead and degraded efficiency.
In contrast, \Sys enables fine-grained distribution of hardware resources based on per-request SLOs and dynamically adjusts both the depth and width of the candidate token tree to adapt to workload changes. This adaptivity allows \Sys to maximally utilize hardware resources, maintaining high efficiency even with large batch sizes. 



As shown in~\Cref{fig:rps_attainment} and~\Cref{fig:rps_goodput}, \Sys's SLO attainment also decreases as the request rate increases. 
This degradation is primarily due to larger batch sizes reducing the average token budget available per request, which limits the effectiveness of speculative decoding.
Additionally, higher request arrival rates introduce higher prefilling overhead, making it increasingly challenging to meet SLOs. 

\paragraph{Changing application distribution.}
In this evaluation, we fix the request arrival rate at 4.0 requests per second and vary the proportion of latency-stringent requests. This setup allows us to evaluate how \Sys performs compared to baseline systems in terms of SLO attainment and goodput under different levels of workload stringency.

As shown in~\Cref{fig:proportion}, \Sys consistently outperforms all baselines across varying proportions of latency-stringent requests.
\Sys maintains stable SLO attainment in all scenarios, while the performance of the baseline systems fluctuates significantly with workload distribution.
\Sys reduces the number of SLO violations by up to 4.3$\times$ and 3.7$\times$ compared to the best-performing baseline under the two model settings, respectively. It also achieves up to 30\% and 64\% higher goodput over the best baseline.

The SLO attainment and goodput of vLLM and Sarathi-Serve drop sharply as the fraction of urgent requests grows. This is because continuous batching systems can only satisfy stringent SLOs with small batch sizes. As the system accumulates more requests, batch sizes grow, increasing latency and causing SLO violations for time-sensitive requests.
In contrast, SD-based methods---vLLM-Spec and \Sys---exhibit the opposite trend. SD accelerates request processing, helping satisfy tighter SLOs even as the share of urgent requests increases. As a result, their SLO attainment remains steady or even improves under higher stringency.

Interestingly, both the SLO attainment and goodput of speculative decoding methods increase as the proportion of urgent requests rises. This is because a lower share of urgent requests corresponds to a higher share of category-3 requests (e.g., summarization) with longer contexts, which increases the prefilling overhead. vLLM-Spec, which lacks awareness of individual decoding speeds, cannot effectively mitigate this overhead. In contrast, \Sys dynamically adapts based on each request’s decoding progress and SLO, enabling smarter compute allocation and improved performance in both SLO attainment and throughput.


\paragraph{Changing SLO-Scale}
In this evaluation, we fix the request rate at 4.0 RPS and set the proportion of urgent requests to 0.6. We then vary the SLO scale of the most urgent request relative to the baseline latency to assess each system’s ability to meet increasingly strict latency requirements.
As shown in~\Cref{fig:slo_scale}, all systems experience reduced SLO attainment and goodput as the SLO scale becomes more stringent. However, \Sys consistently maintains the highest performance across all settings. It achieves up to 4.61$\times$ and 3.05$\times$ lower violation rates, and up to 1.38$\times$ higher goodput than the best baseline across the two evaluated models.
Continuous batching-based systems fail to meet SLOs when the scale drops below 1.0, causing their SLO attainment to fall below 40\%. While vLLM-Spec supports SLO scales below 1.0, it lacks the ability to prioritize urgent requests, leading to lower SLO attainment compared to \Sys.


\subsection{Ablation and Sensitivity Study}
\label{subsec:ablation}
\begin{figure}
    \centering
    \includegraphics[width=0.95\linewidth]{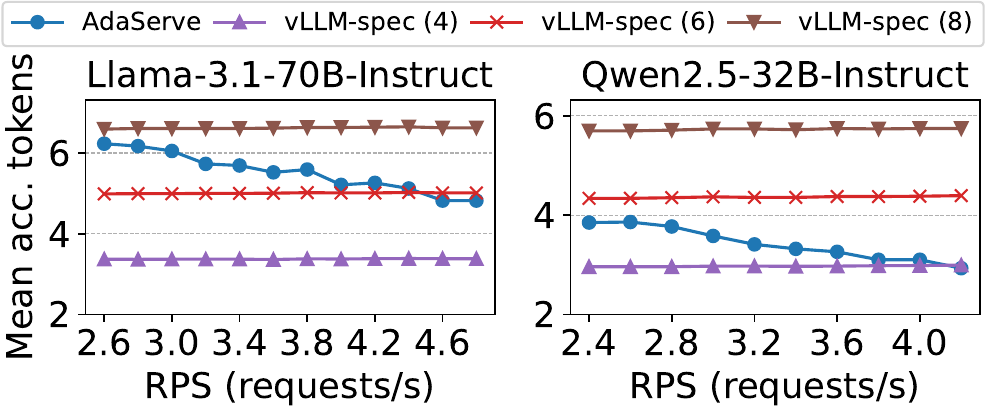}
    \caption{Mean accepted tokens per request per verification in speculative decoding.}
    \label{fig:acc}
\end{figure}


\paragraph{Speculation Accuracy.}
We evaluate the speculation accuracy of \Sys by measuring the average number of tokens accepted by the LLM per verification step per request. As shown in~\Cref{fig:acc}, \Sys achieves high acceptance rates at low RPS levels, which gradually decrease as RPS increases. This behavior aligns with our adaptive strategy for adjusting the depth and width of the candidate tree: when the workload is light, \Sys speculates more aggressively to maximize speedup; under heavy load, it adopts a more conservative approach to reduce verification overhead.
In contrast, vLLM-Spec employs a static speculation strategy, resulting in a constant average acceptance rate regardless of RPS. However, as shown in~\Cref{fig:rps_attainment} and~\Cref{fig:rps_goodput}, this static approach underperforms, particularly at high RPS, demonstrating the effectiveness of \Sys’s dynamic adaptation.

\begin{figure}
    \centering
    \includegraphics[width=0.98\linewidth]{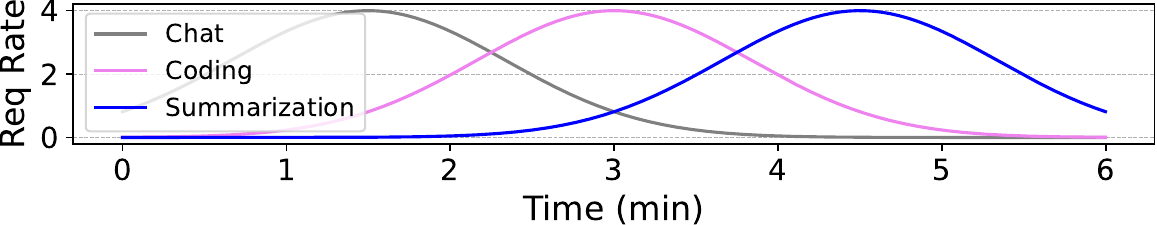}
    \caption{Request arrival pattern of the synthetic trace.}
    \label{fig:fluc_trace}
\end{figure}

\begin{figure}
    \centering
    \includegraphics[width=0.98\linewidth]{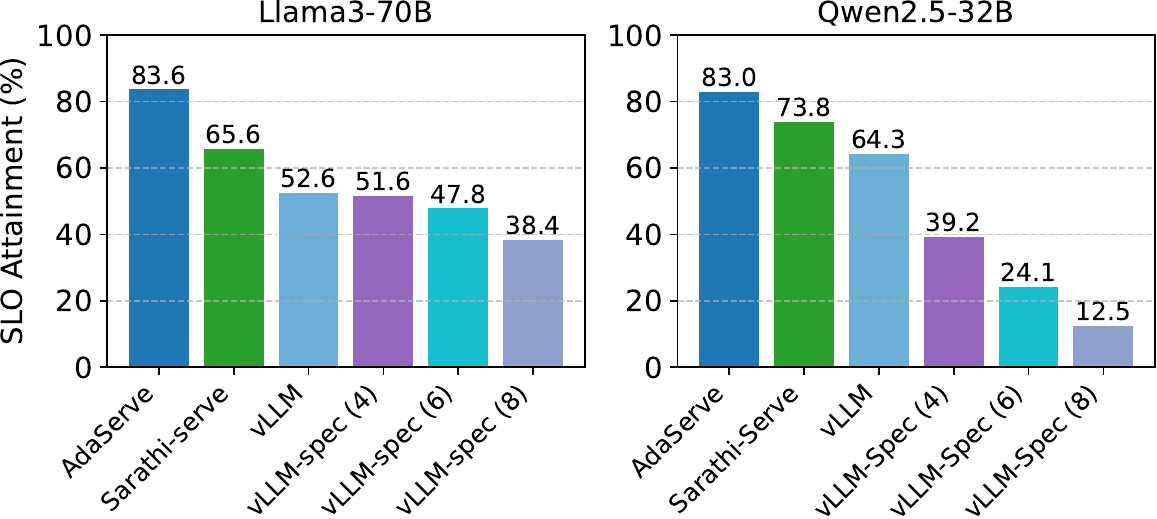}
    \caption{SLO attainment under the synthetic trace.}
    \label{fig:fluc_attainment}
\end{figure}


\paragraph{Sensitivity to Workload Fluctuations.}
We evaluate system performance under workload fluctuations using a synthetic trace where different request categories peak at different times. The request arrival patterns are visualized in~\Cref{fig:fluc_trace}. The SLO attainment is shown in~\Cref{fig:fluc_attainment}. The results highlight the strength of \Sys in handling bursty traffic from individual applications, consistently achieving higher SLO attainment compared to baseline systems.

\begin{figure}
    \centering
    \includegraphics[width=0.98\linewidth]{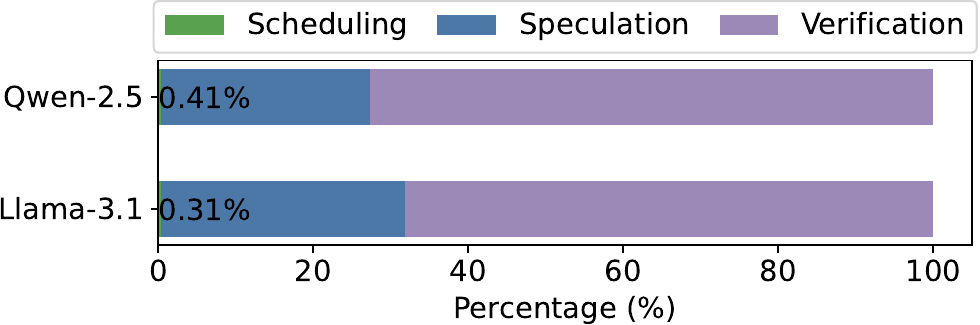}
    \caption{Latency breakdown of \Sys.}
    \label{fig:overhead}
\end{figure}

\paragraph{Latency Breakdown of \Tech.}
We evaluate the runtime overhead of \Tech by measuring the time spent in its three main components: speculation, selection, and verification. Speculation and verification are GPU-intensive, while selection runs on the CPU. Our primary goal is to assess the CPU overhead. As shown in \Cref{fig:overhead}, the CPU overhead is minimal---only 0.41\% and 0.31\% on the two evaluated models---compared to the overall serving time. These results demonstrate that \Tech imposes negligible overhead and is well-suited for integration into speculative decoding-based serving systems.

\section{Related Work}
\label{sec:related}

\paragraph{LLM serving systems}

A wide range of systems have been proposed to enhance the efficiency and scalability of LLM serving~\cite{yu2022orca, vllm, patel2024splitwise, agrawal2024taming, zhong2024distserve, holmes2024deepspeed, tensorrtllm, qin2024mooncake, sglang, sheng2023flexgen, mei2024helix}. Orca~\cite{yu2022orca} introduces continuous batching, allowing new requests to join an ongoing batch without waiting for its completion---a technique now standard in modern serving systems. vLLM~\cite{vllm} identifies GPU memory fragmentation as a key throughput bottleneck and addresses it with PagedAttention, which organizes memory in pages to reduce fragmentation.
Several systems optimize the scheduling of the prefill and decode stages~\cite{patel2024splitwise, zhong2024distserve, agrawal2024taming}. Splitwise~\cite{patel2024splitwise} and DistServe~\cite{zhong2024distserve} observe distinct hardware utilization patterns in these stages and propose executing them on separate nodes to better utilize resources. Sarathi-Serve~\cite{agrawal2024taming}, by contrast, notes that prefill is compute-intensive while decode often underutilizes compute resource, and improves efficiency by co-batching requests from both stages.
Another optimization direction is prefix caching, motivated by prompt repetition in multi-turn interactions~\cite{sglang,qin2024mooncake}. This technique caches KV states of frequently reused prefixes in GPU memory to reduce latency.
These approaches are largely orthogonal and complementary to \Sys, which focuses on multi-SLO LLM serving---an area that remains underexplored in existing systems.

\paragraph{Speculative decoding (SD)}
A variety of algorithms have been proposed to determine the topology of the token tree in SD. Early approaches \cite{miao2023specinfer, medusa, eagle1} use a fixed tree structure for each iteration. More recent methods~\cite{eagle3, suffix_decoding} enable adaptive tree construction. Sequoia \cite{chen2024sequoia} adjusts tree size based on hardware specifications and applies dynamic programming to determine a global tree structure. In contrast, Eagle-2 \cite{li2024eagle} constructs the tree based on input context: the draft model performs beam search to propose a candidate tree and selects the top-$m$ tokens with the highest global acceptance rates.
Unlike prior work, \Sys addresses both tree construction and the fine-grained allocation of hardware resources across requests with diverse needs. It also dynamically adjusts the speculative configuration under varying workloads.

Recent efforts have explored SD in dynamic online serving settings. SmartSpec \cite{liu2024optimizingspecdecoding} adaptively tunes draft sequence lengths based on workload and acceptance rates. SpecServe \cite{huang2025specserve} incorporates service-level objectives (SLOs) into the scheduling process. However, neither supports tree-based decoding or accounts for heterogeneous request demands.
A concurrent work \cite{chen2025slos} addresses the multi-SLO challenge using dynamic programming to schedule SD. In contrast, \Tech in \Sys employs a lower-complexity, tree-based approach that improves performance. To our knowledge, \Sys is the first to address multi-SLO serving using batched, tree-based SD to intelligently allocate compute resources across diverse requests.

\section{Conclusion}

To address the growing demand for serving LLM requests with diverse service-level objectives (SLOs), this paper presents \Sys, the first LLM serving system explicitly designed for multi-SLO serving. We formalize the multi-SLO serving problem and identify key limitations in existing approaches based on continuous batching and conventional speculative decoding. To overcome these challenges, we propose a theoretically optimal algorithm for constructing token trees that balance SLO attainment and system throughput. Building on this foundation, we develop \Tech, a practical and efficient solution that incorporates four stages: speculation, SLO-customized selection, throughput-optimized selection, and verification. We implement \Tech within \Sys and evaluate its performance across a range of multi-SLO workloads. Our results show that \Sys significantly outperforms state-of-the-art LLM serving systems, achieving higher SLO satisfaction and better goodput across diverse application scenarios.

\iffinal
\section*{Acknowledgment}
This research is supported by NSF awards CNS-2147909, CNS-2211882, and CNS-2239351, and research awards from Amazon, Cisco, Google, Meta, NVIDIA, Oracle, Qualcomm, and Samsung.
The views and conclusions contained in this document are those of the authors and should not be interpreted as representing the official policies, either expressed or implied, of any sponsoring institution, the U.S. government or any other entity.
\fi

\bibliographystyle{plain}
\bibliography{reference} 

\clearpage
\appendix
\section{Expected Number of Accepted Tokens}
\label{appendix:exp_acc}

Let  $n_{acc}$  denote the number of accepted tokens in a verification process. Define $p_i$ as the probability of token $i$ being accepted. The average acceptance rate across the $n$  tokens in the verification batch is given by
$p = \frac{\sum_{i=1}^{n} p_i}{n}$.
We can compute the expected number of accepted tokens as follows:

\begin{align}
    E[n_{acc}] &= E[\sum_{i=1}^n \mathbbm{1}(\text{token i is accepted})] \\
    &=\sum_{i=1}^n E[\mathbbm{1}(\text{token i is accepted})] \\
    &= \sum_{i=1}^n p_i \\
    &= np
\end{align}

The acceptance probability $p_i$ decreases exponentially with the depth of token $i$ in the speculation tree. Moreover, for tokens sharing the same parent node in the token tree, their acceptance probabilities sum to 1.
Consequently, given a fixed number of requests in the batch, increasing the number of tokens $n$ in the verification process leads to a lower average acceptance rate $p$.

\section{Proof for Connectivity}
\label{proof:connect}

\begin{proof}
    In this proof, we demonstrate that the output nodes of an iterative greedy algorithm selecting nodes with the highest values on a token tree form a valid tree.

    Language models assign a probability less than 1 to each token given an input token sequence. Therefore, for any node $v$ in the token tree (except for the root node), we have:
   \begin{align*}
       f(v) < f(parent(v))
   \end{align*}
    where $parent(v)$ denotes the parent of node $v$ in the token tree.

    The iterative greedy algorithm ensures that when a node $v$ is selected, all nodes $v'$ with $f(v') > f(v)$ have already been selected, including $parent(v)$. Consequently, when a node is selected, its parent is guaranteed to have been selected beforehand.

    We prove that the selected nodes are connected using induction:
    \begin{enumerate}
        \item \textit{Base Case}: The root node is selected first because it has the highest value ($f(root) = 1 > f(v)$ for all other nodes $v$).
        \item \textit{Inductive Step}: Assume that at step $n-1$, the selected nodes are connected. For a node $v$ at step $n$, the algorithm ensures that $parent(v)$ is selected before $v$, $f(parent(v)) > f(v)$. Thus, $v$ is connected to the selected nodes.
    \end{enumerate}

By induction, all selected nodes form a valid, connected tree.

\end{proof}

\section{Optimality Proof for \Cref{alg:tree_construct_optimal}}
\label{proof:opt}

\begin{proof}
The proof is divided into two main parts:

    \begin{enumerate}
        \item 	If \Cref{alg:tree_construct_optimal} returns INVALID, no feasible solution exists.
        \item 	If a feasible solution exists, the solution returned by \Cref{alg:tree_construct_optimal} is optimal.
    \end{enumerate}

Preliminaries and Notation:

    \begin{itemize}
        \item 	For each request $r_i$, we have a token tree $T_{inf}(r_i)$.
        \item 	Each node $v$ in $T_{inf}(r_i)$ is associated with a path probability $f(v)$.
        \item 	The goal for each request $r_i$ is to achieve a target path probability $A(r_i)$ (the SLO).
        \item 	We have a total budget $B$, which is the maximum number of tokens (nodes) that can be selected across all requests.
        \item 	We define $N_i$ as the minimal number of tokens needed to be selected from $T_{inf}(r_i)$ to achieve $A(r_i)$.
    \end{itemize}

\begin{lemma}[Minimality in Threshold Attainment]
\label{lemma:minimal}
Given a token tree and a threshold $\tau$, consider a greedy algorithm that repeatedly selects the node with the highest $f(v)$ not yet chosen, until the sum of $f(v)$ of the chosen nodes meets or exceeds $\tau$. Suppose this process stops after selecting $n$ nodes. Then there is no subset of fewer than $n$ nodes from the tree whose sum of $f(v)$ is at least $\tau$.
\end{lemma}

Proof of \Cref{lemma:minimal}:
By construction, after selecting $n-1$ nodes, the greedy algorithm did not meet the threshold $\tau$. Therefore, any subset of size less than $n$ cannot meet or exceed $\tau$, since the greedy set of $n-1$ nodes is by definition a best possible subset of that size in terms of cumulative $f(v)$ (no other subset of $n-1$ nodes can have a greater sum than the greedily chosen $n-1$). Thus, $n$ is the minimal number of nodes required to surpass the threshold.

\paragraph{Part 1:} If \Cref{alg:tree_construct_optimal} Returns INVALID, No Feasible Solution Exists

Consider running \Cref{alg:tree_construct_optimal}. For each request $r_i$:

    \begin{enumerate}
        \item The algorithm attempts to meet $A(r_i)$ by repeatedly choosing the highest $f(v)$ node from $T_{inf}(r_i)$ not yet chosen by any request, until $A(r_i)$ is reached or the budget $B$ is exhausted.
        \item If at some step $i$, the algorithm cannot find enough tokens to achieve $A(r_i)$ (i.e., it runs out of budget before $A(r_i)$ is met), it returns INVALID.
    \end{enumerate}

By \Cref{lemma:minimal}, the minimal number of tokens needed to achieve $A(r_i)$ is $N_i$. If the algorithm fails at request $i$, it means it has already allocated tokens to previous requests $r_1, \dots, r_{i-1}$ optimally (since it picks the highest probability nodes first). Thus, by the time it considers $r_i$, it has spent at least $N_1 + N_2 + \cdots + N_{i-1}$ tokens. If it cannot fulfill $A(r_i)$, it implies $N_1 + \cdots + N_i > B$. Therefore, there is no way to allocate $B$ tokens to meet all $A(r_1), \dots, A(r_i)$ simultaneously. Since this reasoning applies for the request where the algorithm fails, if \Cref{alg:tree_construct_optimal} returns INVALID, no feasible solution exists.

\paragraph{Part 2:} If a Feasible Solution Exists, the Returned Solution is Optimal

Now suppose \Cref{alg:tree_construct_optimal} completes successfully. It produces a solution $S$ that satisfies $A(r_i)$ for all $i$ within the budget $B$. We need to show that if there is any other feasible solution $S’$ that also meets all SLOs, then $S$ is at least as good as $S’$ (i.e., $S$ is optimal).

To prove this, we rely on another lemma about the greedy selection of nodes under a fixed budget.

\begin{lemma}[Maximality Under a Fixed Budget]
\label{lemma:maximal}
Given a token tree and a budget $b$, let a greedy algorithm select the top $b$ nodes in terms of $f(v)$ from that tree. This selection maximizes the sum of $f(v)$ over all subsets of size $b$.
\end{lemma}

Proof of \Cref{lemma:maximal}:
Suppose for contradiction that there is a subset $V’$ of size $b$ whose total sum of $f(v)$ is greater than that of the subset $V$ chosen by the greedy algorithm. Since the greedy algorithm picks the top $b$ nodes, every node in $V \setminus V’$ must have $f(v)$ greater than or equal to that of any node in $V’ \setminus V$. By swapping the lower-probability nodes in $V’$ with the higher-probability nodes from $V$, we form a new subset that has a sum at least as large as $V’$. But this new subset is precisely $V$, contradicting the assumption that $V’$ has a strictly greater sum. Thus, $V$ is optimal.

Establishing Optimality of the Returned Solution $S$:

    \begin{enumerate}
        \item 	Define $N_i$ as the minimal number of tokens required to achieve $A(r_i)$ for each request $r_i$. Note that $M_i(S) \ge N_i$ for the solution $S$ returned by the algorithm, where $M_i(S)$ is the number of tokens allocated to $r_i$ in $S$. The same holds for any other feasible solution $S’$: $M_i(S’) \ge N_i$.
        \item 	Suppose there exists a valid solution $S’$ that is better than $S$. Being “better” might mean it uses fewer tokens or achieves a higher sum of $f(v)$ for the given budget. Consider how $S’$ distributes tokens among requests: there must be some difference in the number of tokens allocated to at least one request, otherwise they are identical solutions.
        \item 	Fix a particular distribution of the budget across the requests. For any single token tree $T_{inf}(r_i)$ and a given number of tokens $M_i$, by \Cref{lemma:maximal}, the greedy choice of $M_i$ nodes yields the maximum possible sum of $f(v)$ for that budget on $r_i$. Thus, if $S’$ differs from $S$, but assigns the same number of tokens $M_i(S’)$ to request $r_i$ as $S$ does, then to improve upon $S$‘s solution, $S’$ must choose nodes with a strictly greater total sum of $f(v)$ than $S$ under the same budget $M_i(S)$. This is impossible due to \Cref{lemma:maximal}, since $S$ is constructed by a greedy procedure.
        \item 	Hence, any improvement in one request’s allocation in $S’$ would require changing the budget distribution among requests. However, after ensuring the minimal quotas $N_i$ for each request (which both $S$ and any feasible $S’$ must respect), the second step of the algorithm in $S$ distributes the remaining tokens globally in a greedy manner. This global greedy step ensures that no other distribution of these “extra” tokens can yield a strictly better sum, since that would contradict the global maximality of the greedy choice.
    \end{enumerate}
    
In other words, if $S’$ tries to reallocate tokens among requests (while still meeting all SLOs), any purported improvement can be dismantled by applying \Cref{lemma:maximal} within each token tree. Ultimately, this shows that no $S’$ better than $S$ can exist.

Conclusion:

    \begin{enumerate}
        \item 	If \Cref{alg:tree_construct_optimal} returns INVALID, no feasible solution can exist, since the minimal required tokens to meet the SLOs of the first $i$ requests already exceed $B$.
        \item 	If a feasible solution exists, the solution returned by \Cref{alg:tree_construct_optimal} must be optimal. Any other solution that meets all SLOs cannot be strictly better, due to the maximality properties of the greedy selections both per-request and globally.
    \end{enumerate}

Thus, \Cref{alg:tree_construct_optimal} is correct and optimal.
\end{proof}

\end{document}